\documentclass[journal]{IEEEtran}

\usepackage{cite}

\usepackage{epsfig}
\usepackage{graphicx}
\usepackage{epstopdf} %converting to PDF
\usepackage{caption}

\usepackage{amsmath}
\usepackage{amsfonts}
\usepackage{amssymb}

\begin{document}

\title{Decoupled Spatial Neural Attention for Weakly Supervised Semantic Segmentation}

\author{Tianyi~Zhang,~
        Guosheng~Lin,~
        Jianfei~Cai,~
        Tong~Shen,~
        Chunhua~Shen,~
        Alex~C.~Kot,~
        \thanks{T. Zhang is with the Interdisciplinary Graduate School, Nanyang Technological University (NTU),Singapore 639798 (e-mail:zh0023yi@e.ntu.edu.sg).}
        \thanks{G. Lin and J. Cai are with the School of Computer Science and Engineering, Nanyang Technological University (NTU),Singapore 639798 (e-mail:gslin@ntu.edu.sg; asjfcai@ntu.edu.sg).}
        \thanks{S. Tong and C. Shen are with School of Computer Science, The University of Adelaide, Adelaide, SA 5005, Australia (e-mail:tong.shen@adelaide.edu.au; chunhua.shen@adelaide.edu.au).}
        \thanks{A. C. Kot is with the School of Electrical and Electronic Engineering, Nanyang Technological University (NTU),Singapore 639798 (e-mail:eackot@ntu.edu.sg).}               
        }

\maketitle

\begin{abstract}
Weakly supervised semantic segmentation receives much research attention since it alleviates the need to obtain a large amount of dense pixel-wise ground-truth annotations for the training images.  Compared with other forms of weak supervision, image labels are quite efficient to obtain. 
In our work, we focus on the weakly supervised semantic segmentation with image label annotations. 
Recent progress for this task has been largely dependent on the quality of generated pseudo-annotations. 
In this work, inspired by spatial neural-attention for image captioning, we propose a decoupled spatial neural attention network for generating pseudo-annotations.
Our decoupled attention structure could simultaneously identify the object regions and localize the discriminative parts which generates high-quality pseudo-annotations in one forward path. 
The generated pseudo-annotations lead to the segmentation results which achieve  the state-of-the-art in weakly-supervised semantic segmentation.
\end{abstract}


\IEEEpeerreviewmaketitle

\section{Introduction}
\label{Sec:intro}

Semantic segmentation is a task to assign semantic label to every pixel within an image.
In recent years, Deep convolutional neural networks (DCNNs) \cite{long2015fully,chen2014semantic,lin2016refinenet} have brought  great improvement in semantic segmentation performance. Training DCNNs  in a fully-supervised setting with pixel-wise ground-truth annotation achieves state-of-the-art semantic segmentation accuracy.  However, the main limitation of such fully-supervised setting is that it is labor-intensive to obtain a large amount of accurate pixel-level annotations for training images. On the other hand, datasets with only image-level annotations are much easier to obtain. Therefore, weakly-supervised semantic segmentation supervised only with image labels has received much attention.

The performance of weakly supervised semantic segmentation with image-level annotation has been remarkably improved by introducing efficient localization cues \cite{kolesnikov2016seed,oh2017exploiting,wei2017object}. The most widely used pipeline in weakly supervised semantic segmentation is to first estimate pseudo-annotations for the training images based on localization cues and then utilize the pseudo-annotations as the ground-truth to train the segmentation DCNNs. 
Clearly the quality of pseudo-annotations directly affects the final segmentation performance. In our work, we follow the same pipeline 
and mainly focus on the first step, which is to generate high-quality pseudo-annotations for the training images with only image-level labels.
In recent years, top-down neural saliency \cite{zhou2016learning,zhang2016top,selvaraju2016grad} performs well in weakly-supervised localization tasks and consequently has been widely applied in generating pseudo-annotations  for semantic segmentation supervised with image-level labels. 
However, as is pointed out by previous works \cite{wei2017object}, such top-down neural saliency is good at identifying the most discriminative regions of the objects instead of the whole extent of the objects. Thus the pseudo-annotations generated by these methods are far from the ground-truth annotations.
To alleviate this problem, some works  consist of multiple ad-hoc processing steps (\textit{e.g.}, iterative training), which are difficult to implement. Some works introduce external information (\textit{e.g.}, web data) to guide the supervision, which greatly increase data and computation load. On the contrary, our work proposes a principal pipeline which is simple and effective to implement.

\begin{figure}[t]
\begin{center}
   \includegraphics[width=1.0 \linewidth]{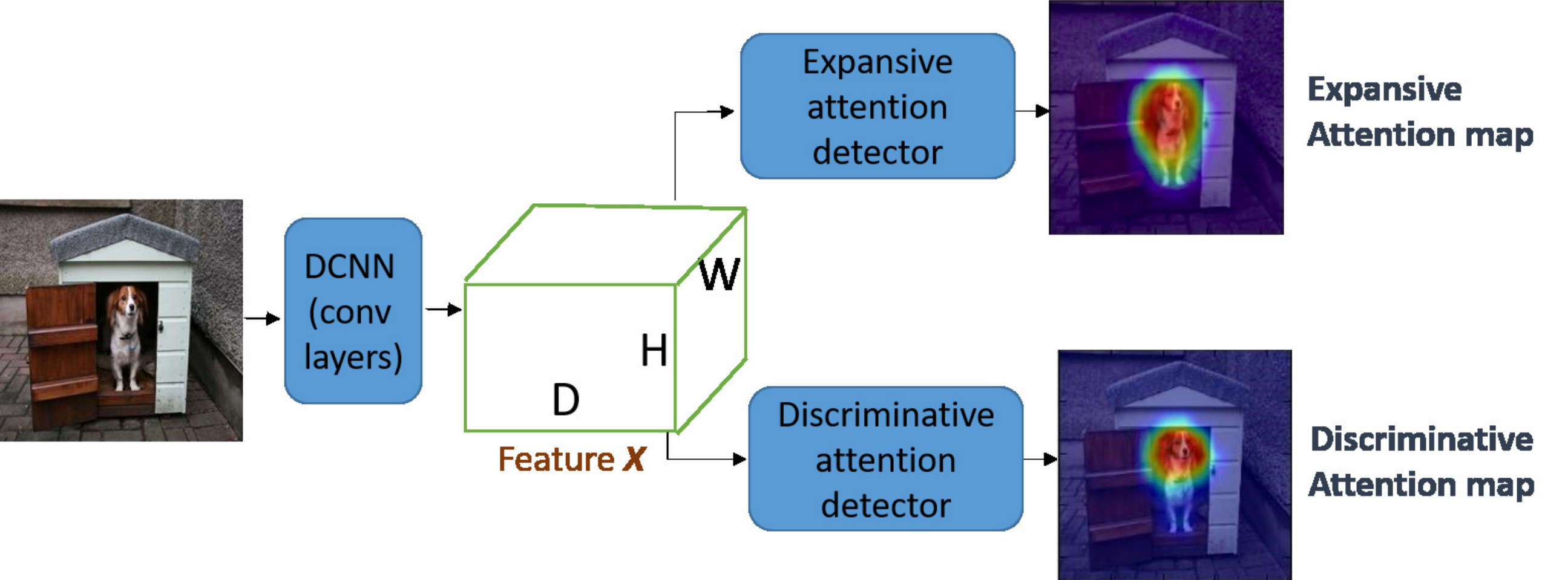}
\end{center}
   \caption{The brief introduction of our proposed decoupled attention structure. It simultaneously generates two attention maps, namely Expansive attention map and Discriminative attention map. The Expansive attention map is to identify the object regions while the Discriminative attention map is to mine the discriminative parts. }

\label{fig:concise_dual_atten}
\end{figure}

Our aim is to generate pseudo-annotations for weakly supervised semantic segmentation efficiently and  effectively. Inspired by the spatial neural attention mechanism which has been widely used in VQA \cite{yang2016stacked} and image captioning \cite{xu2015show}, we introduce spatial neural attention into our pseudo-annotation generation pipeline and propose a decoupled spatial neural attention structure 
which simultaneously localizes the discriminative parts and estimates object regions in one end-to-end framework. 
Such structure helps to generate effective pseudo-annotations in one forward pass. The brief description of our decoupled attention structure is illustrated in Fig.~\ref{fig:concise_dual_atten}.

Our major contributions can be summarized as follows:
\begin{itemize}
\item We introduce spatial neural attention and propose a decoupled attention structure for generating pseudo-annotations for weakly supervised semantic segmentation.

\item Our decoupled attention model outputs two attention maps which focus on identifying object regions and mining the discriminative parts respectively. These two attention maps are complimentary to each other to generate high-quality pseudo-annotations.

\item We employ a simple and effective pipeline without heuristic multi-step iterative training steps,  which is different from most existing methods of weakly-supervised semantic segmentation.

\item We perform detailed ablation experiments to verify the effectiveness of our decoupled attention structure. We achieve state-of-the-art weakly supervised semantic segmentation results on Pascal VOC 2012 image segmentation benchmark.

\end{itemize}

\section{Related Work}

\subsection{Weakly Supervised Semantic Segmentation}
In recent years, the performance of semantic segmentation has been greatly improved with the help of Deep Convolutional Neural Networks (DCNNs) \cite{long2015fully,noh2015learning,lin2016refinenet,shen2017learning,
zhao2017pyramid,chen2016attention,zheng2015conditional,li2017not,lin2016efficient}. Training DCNNs for semantic segmentation in a fully-supervised pipeline requires pixel-wise ground-truth annotations, which is very time-consuming to obtain. 

Thus weakly-supervised semantic segmentation receives research attention to alleviate the workload of pixel-wise annotation for the training data. Among the weakly-supervised settings,  image-level labels are the easiest annotations to obtain. As for the semantic segmentation with image-level labels, some early works \cite{pathak2014fully,pinheiro2015image} tackle this problem as multiple instance learning (MIL) problem which views each image being positive if at least one pixel/superpixel within it is positive, and negative if all of the pixels are negative. Other early works \cite{papandreou2015weakly} apply Expectation-Maximization (EM) procedure, which alternates between predicting pixel labels  and optimizing DCNNs parameters. However, due to the lack of effective location cues, the performance of early works is not satisfactory enough.

The performance of semantic segmentation with image-level labels was significantly improved after introducing location information to generate localization seeds/pseudo-annotations for segmentation DCNNs. The quality of pseudo-annotations directly influences the segmentation results. There are several categories of methods to estimate pseudo-annotations. The first category is Simple-to-Complex (STC) strategy \cite{wei2015stc,hou2016mining,shen2017weakly}. The methods in this category assume that the pseudo-annotations of simple images (\textit{e.g.}, web images) can be accurately estimated by saliency detection \cite{wei2015stc} or co-segmentation \cite{shen2017weakly}. Then the segmentation models trained on the simple images are utilized to generate  pseudo-annotations for the complex images. The methods in this category usually require a large amount of external data which consequently increase the data and computation load.
The second category is region-mining based methods. 
The methods in this category rely on region-mining  methods \cite{zhou2016learning,selvaraju2016grad,zhang2016top} to generate discriminative regions as  localization seeds.  
Since such localization seeds mainly sparsely lie in the discriminate parts instead of the whole extent of the objects, which is far from the ground-truth annotation, many works try to alleviate this problem by expanding the localization seeds to the size of objects.
Kolesnikov et al. \cite{kolesnikov2016seed} expand the seeds by aggregating the pixel-scores by global weighted rank-pooling. Wei et al. \cite{wei2017object} apply an adversarial-erasing approach which iterates between suppressing the most discriminative image region and training the region mining model. It gradually localizes the next most discriminative regions through multiple iterations and  merges all the mined discriminative regions into final pseudo-annotations. Similarly Two-phase \cite{kim2017two} captures the full extent of the objects by suppressing and mining processing in two phases.
 Some works \cite{oh2017exploiting, hou2016mining} utilize external dependencies such as fully-supervised saliency method \cite{liu2016dhsnet} trained on additional saliency datasets to facilitate estimating object scales. 
 
To generate high-quality pseudo-annotations, the first category focuses on the quality of training data while the second category focuses on post-processing the localization seeds which is independent of the region-mining model structure. Different from previous methods, we focus on designing a region-mining model structure which is  likely to highlight the object region.  We aim to generate pseudo-annotations for  weakly-supervised semantic segmentation in a single forward path without external data or external prior for efficiency and simplicity purpose.

\subsection{Mining Discriminative Regions}

 In this section we introduce some region-mining methods, which have been widely used in generating pseudo-annotations for semantic segmentation with image-level labels.
Recent works of top-down neural saliency \cite{zhou2016learning,selvaraju2016grad,zhang2016top} perform well in weakly supervised localization tasks. Such works identify the discriminative regions respect to each individual class based on image classification DCNNs. Zhang et al.  \cite{zhang2016top} propose  Excitation Backprop to back-propagate in the network hierarchy to identify the discriminative regions.
Zhou et al. \cite{zhou2016learning} propose a technique called Class Activation Mapping (CAM) for identifying discriminative regions  by  replacing fully-connected layers in image classification CNNs with convolutional layers and global average pooling.  Grad-CAM \cite{selvaraju2016grad} is a strictly generalization of CAM \cite{zhou2016learning} without the need of modifying DCNN structure.
 Among the methods listed above, CAM \cite{zhou2016learning} is the most widely used one in weakly-supervised  semantic segmentation \cite{kolesnikov2016seed,kim2017two,wei2017object} for generating pseudo-annotations.

\begin{figure*}[!t]
\begin{center}
   \includegraphics[width= 0.7\linewidth]{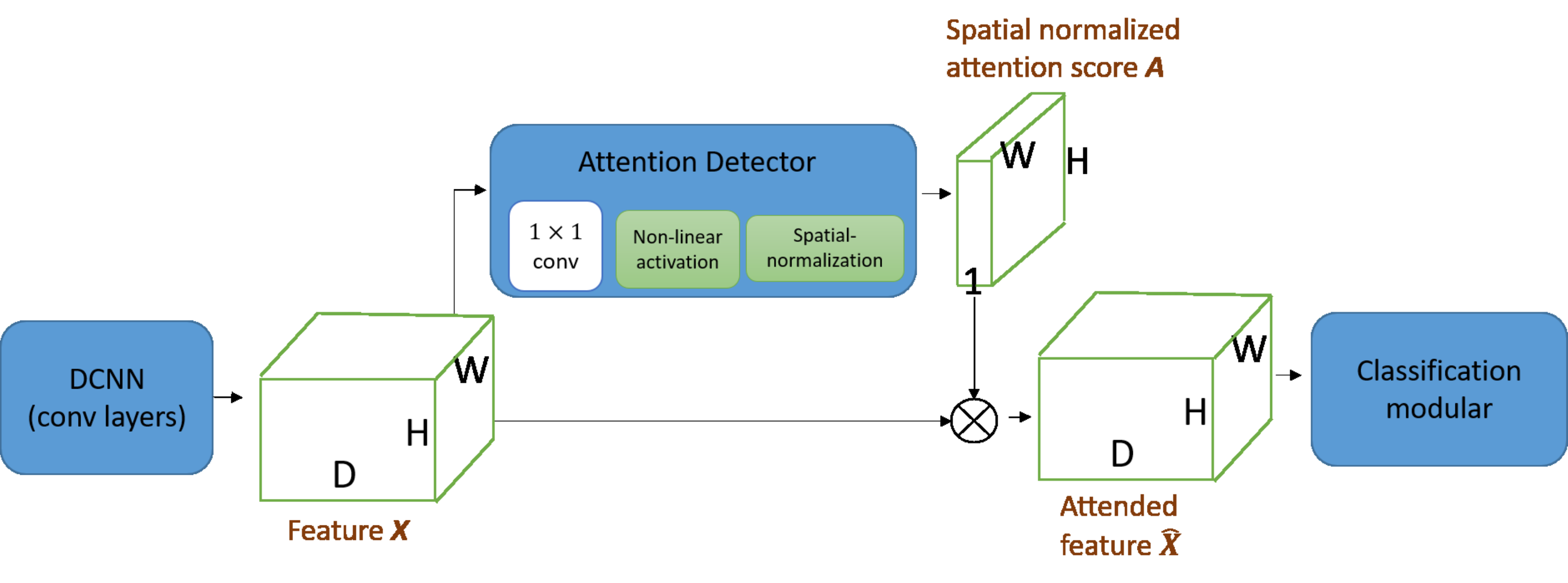}
\end{center}
   \caption{The illustration of the structure of conventional spatial neural attention model. 
   }
\label{fig:coventional_atten}
\end{figure*}

\subsection{Spatial Attention Mechanism}
\label{rela:spa_atten_mech}

Spatial neural attention is a mechanism to assign different weights to different feature spatial regions depending on their feature content. It automatically predicts the weighted heat map  to enhance the relevant features and block the irrelevant features during the training process for specific tasks.  Intuitively, such weighted heat map could be applied to our purpose of  pseudo-annotation generation. 

Spatial neural attention mechanism has been proven to be beneficial in many tasks such as image captioning \cite{you2016image,xu2015show}, machine translation \cite{bahdanau2014neural}, multi-label classification \cite{zhu2017learning}, human pose estimation \cite{chu2017multi} and saliency detection \cite{kuen2016recurrent}. 
Different from the previous works, we are the first to apply the attention mechanism to weakly supervised semantic segmentation to the best of our knowledge.

\section{Approach}
\label{Sec:approrach}

First we give a brief review to the conventional spatial neural attention model in Sec.~\ref{subsec:conv_atten}.
Second we introduce our decoupled attention structure in Sec.~\ref{subsec:dual_path_atten}.
Then we introduce how to further generate pseudo-annotations based on this decoupled attention model in Sec.~\ref{subsec:mask_gener}. Finally, the generated pseudo-annotations are utilized as the ground-truth annotation to train segmentation DCNN networks.

\subsection{Conventional Spatial Neural Attention}
\label{subsec:conv_atten}

In this section, we will give a brief introduction to the conventional spatial neural attention mechanism. 
The conventional spatial neural attention structure is illustrated in Fig.~\ref{fig:coventional_atten}. 
The attention structure consists of two modular branches: the main branch modular and attention detector modular.
The attention detector modular is jointly trained  with the main stream modular end-to-end.

Formally we denote the output features of some convolutional/pooing  layers by $\textbf{X} \in \mathbb{R}^{W \times H \times D}$.
The attention detector takes feature map $\textbf{X} \in \mathbb{R}^{W \times H \times D}$ as the input and outputs spatial-normalized attention weights map $\textbf{A} \in \mathbb{R}^{W \times H }$. 
$\textbf{A}$ is applied on $\textbf{X}$ to get attended feature  $\hat{\textbf{X}}\in \mathbb{R}^{W \times H \times D}$.  $\hat{\textbf{X}}$ is fed into the classification modular to output image classification score vector $\textbf{p} \in \mathbb{R}^C$, where $C$ is the number of classes. 

For notation simplicity, we denote the 3-D feature output of DCNNs with upper-case letters and  feature at one spatial location with its corresponding lower-case letters.   For example, $\textbf{x}_{ij} \in \mathbb{R}^{D}$ is the feature at the position $(i,j)$ of $\textbf{X}$.
Attention detector outputs attention weights map $\textbf{A}$  which acts as a spatial regularizer to enhance the relevant regions and suppress the non-relevant regions for feature $\textbf{X}$. Thus we are motivated to utilize the output of attention detector to generate pseudo-annotations for the task of semantic segmentation with image-level labels.
The details of the attention detector modular are as follows:
Attention detector modular consists of a convolutional layer, a non-linear activation layer (Eq.~\eqref{eq:non-linear_act}) and   a spatial-normalization  (Eq.~\eqref{eq:spa_norm}) as is shown in Fig.~\ref{fig:coventional_atten}:
 \begin{equation}
 \label{eq:non-linear_act}
 z_{ij} = \mathrm{F}(\textbf{w}^\intercal\bf{x}_{ij}+b),
 \end{equation}
\begin{equation}
\label{eq:spa_norm}
a_{ij} = \dfrac{z_{ij}}{\sum_{i,j}z_{ij}},
\end{equation}
where $\mathrm{F}$ is a non-linear activation function, such as exponential function in \cite{xu2015show,chu2017multi}. $\textbf{w} \in \mathbb{R}^{D}$ and $b \in \mathbb{R}^{1}$ are the parameters of the attention detector model, which is a $1 \times 1$ convolution layer. 
The attended feature $\hat{\textbf{x}}_{ij} \in \mathbb{R}^D$ is calculated as 
\begin{equation}
\hat{\textbf{x}}_{ij} = \textbf{x}_{ij}a_{ij},
\end{equation}

The classification modular consists of a spatial average pooling layer and $1\times1$ convolutional layer as the image classifier.
We denote the $\textbf{v}_c$ and $h_c$ as the parameters of the classifier for class $c$, thus the classification score for class $c$ is calculated as:
\begin{equation}
\label{eq:class_score}
 p_c =   (\sum_{i,j} \textbf{x}_{ij}a_{ij})^\intercal \textbf{v}_c + h_c,
\end{equation}
where $p_c$ is the score for $c$-th class.

\subsection{Decoupled Attention Structure}
\label{subsec:dual_path_atten}

\begin{figure*}[!t]
\begin{center}
   \includegraphics[width=0.8 \linewidth]{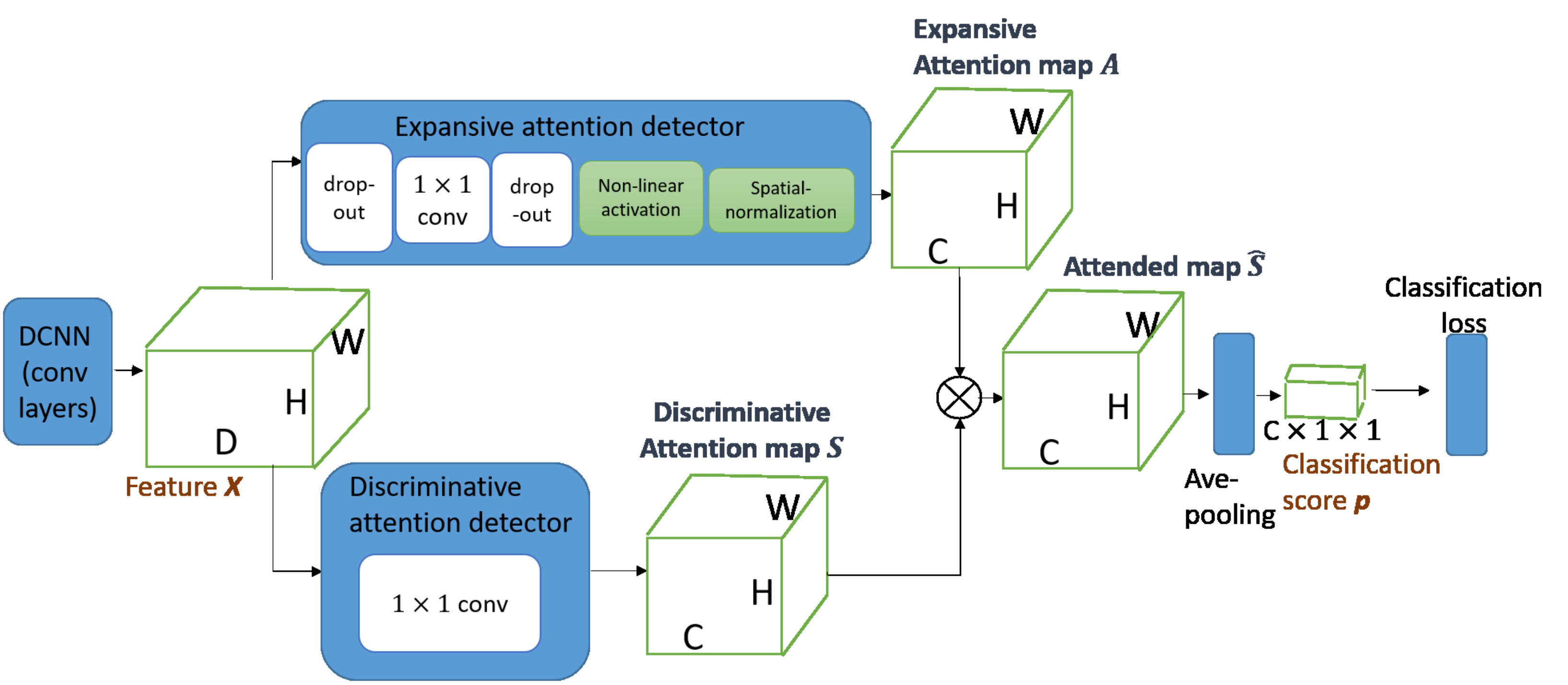}
\end{center}
   \caption{Our proposed decoupled attention structure. The heat map generated by Expansive attention detector is named  \textit{Expansive attention map}, which identifies the  object regions. The heat map generated by the Discriminative attention detector is named  \textit{Discriminative attention map}, which localizes the discriminative parts. }
   
\label{fig:dual_atten}
\end{figure*}

As described in Sec.~\ref{subsec:conv_atten}, the output of conventional spatial neural attention detector is class-agnostic. However, in semantic segmentation task each image may contain objects of multi-classes. Thus  conventional class-agnostic attention map is not applicable to generate pseudo-annotations for such multi-class case since we need to predict pixel-wise label for each semantic class instead of just foreground/background. On the other hand, the output of conventional spatial neural attention detector is aimed to assist the task of image classification  which may not necessarily generate desired pseudo-annotations for weakly-supervised semantic segmentation. Inspired by \cite{zhu2017learning}, we propose our decoupled attention structure especially for the task of generating pseudo-annotations for weakly supervised semantic segmentation to alleviate these problems.

We illustrate our structure in Fig.~\ref{fig:dual_atten}. Such structure  extend the conventional attention structure to multi-class cases. Moreover, it generates two different types of attention maps which identifies object regions and predicts the discriminative parts respectively. 
In Fig.~\ref{fig:dual_atten}, the attention map $\textbf{A} \in \mathbb{R}^{W \times H \times C}$ generated by the Expansive attention detector in the top branch is named \textbf{Expansive attention map}, which aims at identifying object regions. 
The attention map $\textbf{S} \in \mathbb{R}^{W \times H \times C}$ generated by the Discriminative attention detector in the bottom branch is named \textbf{Discriminative attention map}, which aims at mining the discriminative parts. 
Such two attention maps have different properties which are complimentary to each other  to generate pseudo-annotations for weakly supervised semantic segmentation.

The details of the structure in Fig.~\ref{fig:dual_atten} are described as follows:

Expansive Attention detector (E-A detector) modular consists of a convolutional layer, a non-linear activation layer (Eq.~\eqref{eq:non-linear_act_cls}) and   a spatial-normalization step (Eq.~\eqref{eq:spat_norm_cls}):
\begin{equation}
\label{eq:non-linear_act_cls}
z^c_{ij} = \mathrm{F}({\textbf{w}_c}^\intercal  \textbf{x}_{ij}+b^c),
\end{equation}
\begin{equation}
\label{eq:spat_norm_cls}
a^c_{ij} = \dfrac{z^c_{ij}}{\sum_{i,j}z^c_{ij}},
\end{equation}
where the superscript/subscript $c$ denotes the value of $c$-th channel/class.

As we mentioned in Sec.~\ref{Sec:intro}, for the task of pseudo-annotation generation we aim to estimate the whole range of objects instead of only obtaining the most discriminative parts. Thus we design the E-A detector  details as follows:
we set 
\begin{equation}
\mathrm{F}(x)= \log(1+\exp(x))+\epsilon, 
\end{equation}
where $\epsilon=0.1$, similar to \cite{zhuang2016attend}. 
Besides the $1 \times 1$ convolutional layer in the attention detector, we add one drop-out layer right before and after it respectively. Such drop-out layers randomly zero-out the elements in the training features and consequently the attention detector will highlight more relevant features instead of just the most relevant one for successful classification.

The Discriminative attention detector (D-A detector) consists of $1\times1$ convolutional layer whose parameters are denoted as $\textbf{v}_c$ and $h_c$ same as the classifier in Sec.~\ref{subsec:conv_atten}. The D-A detector takes feature map $\textbf{X}$ as the input and outputs class-specific attention map  $\textbf{S} \in \mathbb{R}^{W \times H \times C}$.

The attended feature $\hat{\textbf{S}} \in \mathbb{R}^{W \times H \times C}$ is calculated as:
\begin{equation}
\hat{s}_{ij}^c = s_{ij}^ca_{ij}^c,
\end{equation}
$\hat{\textbf{S}}$ is fed into a spatial average pooling layer to generate image classification score $\textbf{p} \in \mathbb{R}^C$.
Hence $p_c$ is calculated as:
\begin{equation}
\label{eq:class_score_muls}
 p_c =    \sum_{i,j}( \textbf{x}_{ij}^\intercal \textbf{v}_c+ h_c) a^c_{ij}.
\end{equation}
Compared with Eq.~\eqref{eq:class_score}, in our decoupled model the D-A detector is to predict the class score for dense pixel position instead of predicting image label score. Thus our model remains the spatial information for the classification map which is more suitable for semantic segmentation tasks.
The multi-label classification loss for each image is formulated as: 
\begin{equation}
\label{eq:class-loss}
\small
L_{oss} = -  \sum_c \Big[ y_c \log(\dfrac{1}{1+e^{-p_c}})+ (1-y_c)\log(\dfrac{e^{-p_c}}{1+e^{-p_c}}) \Big],
\end{equation} 
where $y_c$ is the binary image label corresponding to $c$-th class.

\begin{figure*}[htb]

\begin{minipage}[b]{0.11\linewidth}
 \centering
  \centerline{\includegraphics[width=2.1cm]{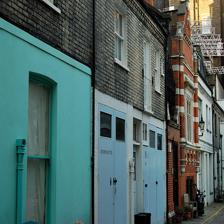}}
 \centerline{\includegraphics[width=2.1cm]{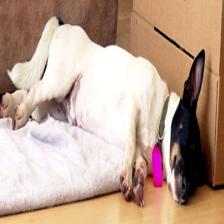}}
\centerline{input image}
\centerline{\qquad}
\end{minipage}
 \hfill
\begin{minipage}[b]{0.11\linewidth}
 \centering
 \centerline{\includegraphics[width=2.1cm]{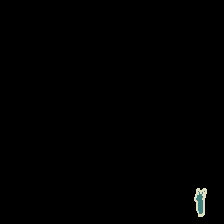}}
 \centerline{\includegraphics[width=2.1cm]{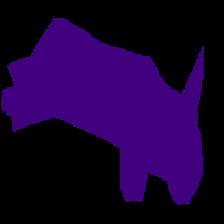}}
\centerline{ground-truth}
\centerline{\qquad}
\end{minipage} 
\hfill
\begin{minipage}[b]{0.11\linewidth}
 \centering
 \centerline{\includegraphics[width=2.1cm]{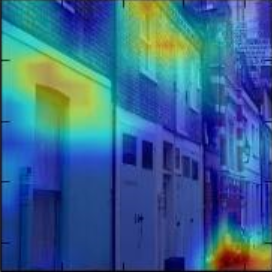}}
 \centerline{\includegraphics[width=2.1cm]{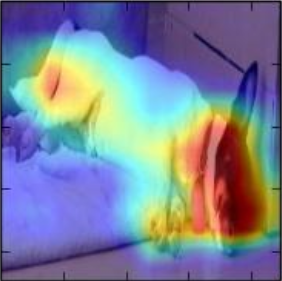}}
\centerline{Expansive }
\centerline{attention map}

\end{minipage}
 \hfill
\begin{minipage}[b]{0.11\linewidth}
 \centering
  \centerline{\includegraphics[width=2.1cm]{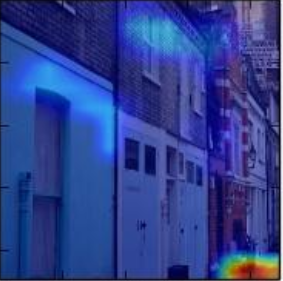}}
 \centerline{\includegraphics[width=2.1cm]{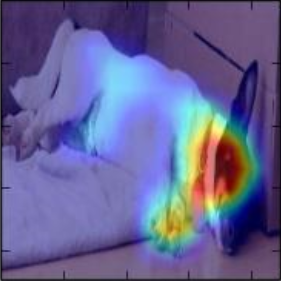}}
\centerline{Discriminative }
\centerline{attention map }
\end{minipage} 
%\vspace{2pt}
\hfill\vline\hfill
\begin{minipage}[b]{0.11\linewidth}
 \centering
  \centerline{\includegraphics[width=2.1cm]{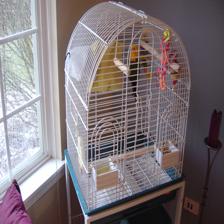}}
 \centerline{\includegraphics[width=2.1cm]{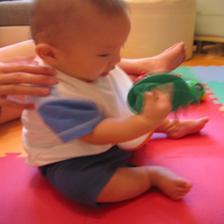}}
\centerline{input image}
\centerline{\qquad}
\end{minipage}
 \hfill
\begin{minipage}[b]{0.11\linewidth}
 \centering
 \centerline{\includegraphics[width=2.1cm]{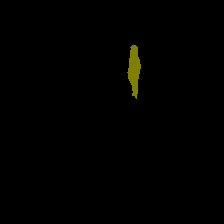}}
 \centerline{\includegraphics[width=2.1cm]{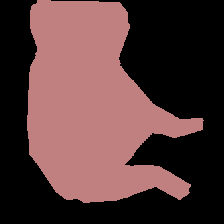}}
\centerline{ground-truth}
\centerline{\qquad}
\end{minipage} 
\hfill
\begin{minipage}[b]{0.11\linewidth}
 \centering
 \centerline{\includegraphics[width=2.1cm]{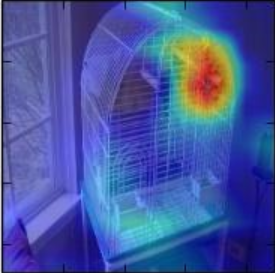}}
 \centerline{\includegraphics[width=2.1cm]{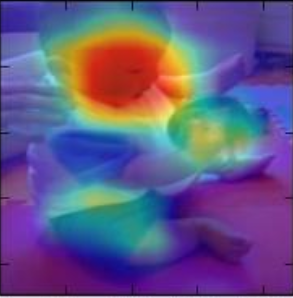}}
\centerline{Expansive }
\centerline{attention map}

\end{minipage}
 \hfill
\begin{minipage}[b]{0.11\linewidth}
 \centering
  \centerline{\includegraphics[width=2.1cm]{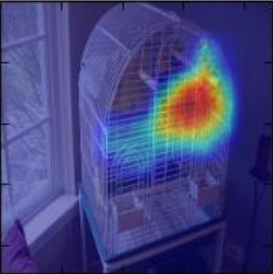}}
 \centerline{\includegraphics[width=2.1cm]{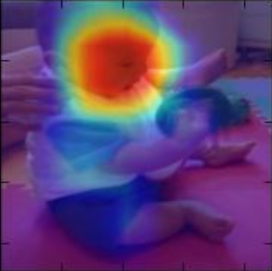}}
\centerline{Discriminative }
\centerline{attention map }
\end{minipage}

\caption{The examples of Expansive attention map and Discriminative attention map.  Expansive attention map and Discriminative attention map have different properties which are suitable for different situations.}
\label{fig:example}
\end{figure*}

We show examples of Expansive attention map  and Discriminative attention map  in Fig.~\ref{fig:example}, which illustrates that 
Expansive attention maps perform well at identifying large object regions while   Discriminative attention maps perform well at mining the small discriminative parts. Such two attention maps show different and complimentary properties.
 Thus we merge such two attention maps using  Eq.~\eqref{eq:merge}
\begin{equation}
\label{eq:merge}
\textbf{T}_c = \hat{p}_c*\textbf{A}_c + (1-\hat{p}_c)*\textbf{S}_c,
\end{equation}
 where $\textbf{A}_c$ is the normalized Expansive attention map and $\textbf{S}_c$ is the normalized Discriminative attention map corresponding to $c$-th class. $\textbf{T}_c$ is the result merged attention map. $\hat{\textbf{p}}$ is the softmax normalization result of the image classification score $\textbf{p}$.
 Such weighted combination is intuitive: the small prediction score usually correspond to the  difficult objects of small size so more weights should be put on the task of mining the discriminative regions.

\subsection{Pseudo-annotation Generation}
\label{subsec:mask_gener}
In this section we describe how to generate pseudo-annotations according to the  attention maps.
We generate pseudo-annotations by simple thresholding following the similar practice used in \cite{kolesnikov2016seed,wei2017object,zhou2016learning}.
Given an attention map of an image with image label set $\mathcal{L}$ (excluding the background),  for each class $c$ in $\mathcal{L}$, we generate foreground regions as follows: first we perform min-max spatial normalization on the attention map corresponding to class $c$ to [0,1] range. Then we threshold the normalized attention map by $>$0.2.
Inspired by \cite{saleh2016built},  we sum the feature map $\textbf{X}$ in the channel dimension and then perform min-max spatial normalization on it. Then we threshold this normalized  map by $<$0.3 to generate background regions.
  Since the regions are generated independently for each class, there  may exist label conflicts at some pixel positions. 
We choose foreground label with smallest region size for the conflict regions following the practice of \cite{kolesnikov2016seed}. We assign the unclassified pixels with void label denoted as $void$ which represents the label is unclear at this position and will not be considered in the training process. 

The generated hard annotations $M$ are coarse and have much unclear area. We can further apply denseCRF \cite{krahenbuhl2011efficient}  on $M$ to generate refined annotations. 
We first describe how to generate the class probability vector for each spatial locations.  
For an image $I$, the labels which are present in $I$ are denoted as $\mathcal{C}_I$. All the class labels in the target dataset is denoted as $\mathcal{C}$. $m_i$ is the mask label of $M$ for pixel $i$. We calculate the class probability vector $z_{i,c}$ for class $c \in \mathcal{C} $ at pixel $i$ as follows:

If $m_i=void$ ,
\begin{equation}
  z_{i,c}= 
\begin{cases}
    \dfrac{1}{|\mathcal{C}_I|}, & \text{if $c \in \mathcal{C}_I $ }\\
    0, & \text{otherwise}
\end{cases}
\end{equation}

Else:

\begin{equation}
  z_{i,c}= 
\begin{cases}
    \tau, & \text{if $c= m_i $ }\\
    \dfrac{1-\tau}{|\mathcal{C}_I|-1}, & \text{if  $c \neq m_i$ and $c \in \mathcal{C}_I $ }\\
    0, & \text{otherwise}
\end{cases}
\end{equation}
where $\tau \in (0.5,1)$ is the manually fixed parameter. $|\mathcal{C}_I|$ represents the number of labels that are present in image $I$. The unary potential is calculated for the probability vector. We apply denseCRF \cite{krahenbuhl2011efficient} with the unary potential and take the result mask as the refined annotations.

\section{Experiments}

\subsection{Experimental Set-up}

\textbf{Dataset and Evaluation Metric}
We evaluate our method on the PASCAL VOC 2012 image segmentation benchmark \cite{everingham2010pascal}, which has 21 semantic classes including background. The original dataset has 1464 training images (\textit{train}), 1449 validation images (\textit{val}) and 1456 testing images (\textit{test}). The dataset is augmented to 10582 training images by \cite{hariharan2011semantic} following the common practice. In our experiments, we only utilize image-level class labels of the training images. We use the \textit{val} images to evaluate our method. As for the evaluation measure for segmentation performance, we use the standard PASCAL VOC 2012 segmentation metric: mean intersection-over-union (mIoU).

\textbf{Training/Testing Settings}
We train the proposed decoupled attention network for pseudo-annotation estimation.
Based on the generated pseudo-annotations we train the state-of-the-art semantic segmentation network to predict the final segmentation result.

We build the proposed decoupled attention model based on $VGG\textrm{-}16$ model. 
We transfer the layers from $VGG\textrm{-}16$ from the first layer to layer $relu5\_3$ as the starting convolutional layers (as is shown in Fig.~\ref{fig:dual_atten}) which outputs $\textbf{X} \in \mathbb{R}^{14 \times 14 \times 512}$.
We use a mini-batch size of 15 images with the data augmentation methods such as random flip and random scale. 
We set the 0.01 as initial learning rate for the $VGG\textrm{-}16$ transferred layers  and 0.1 as the initial learning rate for the attention detector layers. We decrease the learning rate by a factor of 10 after 10 epochs. Training terminates after 20 epochs. 

We train the DeepLab-LargeFOV ($VGG\textrm{-}16$ based) of \cite{chen2014semantic} as our segmentation model.
The input image crops for the network are of size 321$\times$321 and outputs segmentation mask are of size 41$\times$41.
The initial base learning rate is 0.001  and it is decreased by a factor of 10 after 8 epochs. Training terminates after 12 epochs. We use the public available pytorch implementation of Deeplab-largeFOV \footnote[1]{https://github.com/BardOfCodes/pytorch$\_$deeplab$\_$large$\_$fov}. In the inference phase of segmentation, we use multi-scale  inference to combine output score at different scales, which is common practice as in \cite{peng2017large,lin2016refinenet}. The final outputs are post-processed by denseCRF \cite{krahenbuhl2011efficient}.

\begin{table}[!htb]
\begin{center}
\caption{Quantitative evaluation of attention maps.  $mS_{prec}$ is the criteria about localizing the concentrated interior object parts, $mS_{rec}$ is the criteria about covering over the object regions and $mS_{IoU}$  is the criteria for identifying the whole object regions. The results show that Expansive attention map (expan-atten) performs well at covering over and identifying the whole object regions and Discriminative attention map (disc-atten) performs well at localizing  interior partial object regions.
The merged attention map (merged-atten) achieves the highest $mS_{IoU}$, which means it performs best at identifying the whole object regions.}
\label{table:num}
\resizebox{0.9\linewidth}{!}{
\begin{tabular}{|l|c|c|c|}
\hline
 & expan-atten & disc-atten & merged-atten \\
\hline
 $mS_{prec}$ & 52.0 & \textbf{61.2} & $-$\\
 
 $mS_{rec}$ & \textbf{71.9} & 54.4 & $-$\\

 $mS_{IoU}$  & 42.1 & 39.4 & \textbf{44.0}\\
\hline
\end{tabular}
}
\end{center}

\end{table}

\subsection{Ablation Study of Decoupled Attention Model}
\label{Exp:dual_atten}

\subsubsection{Properties of attention maps}

\begin{figure*}[!htb]
\centering

\begin{minipage}[t]{2.6cm}
  \centering
  {\includegraphics[width=2.4cm]{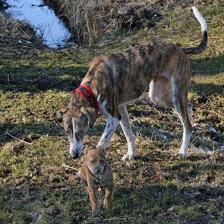}}\\

    {\includegraphics[width=2.4cm]{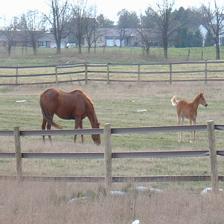} }\\

  {\includegraphics[width=2.4cm]{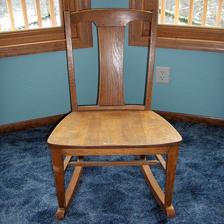} }\\

  {\includegraphics[width=2.4cm]{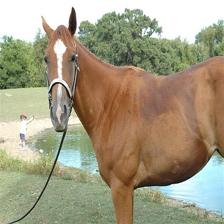} }\\
  
    {\includegraphics[width=2.4cm]{figures/JPEGImages_select/2008_000141.jpg} }\\
    
     {\includegraphics[width=2.4cm]{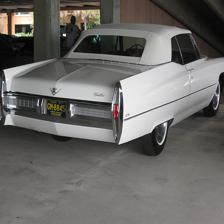} }\\

     {\includegraphics[width=2.4cm]{figures/JPEGImages_select/2008_000939.jpg} }\\

    \centerline{\textbf{raw image}}

\end{minipage}
%atten*******************************************************
\begin{minipage}[t]{2.6cm}
  \centering
  {\includegraphics[width=2.4cm]{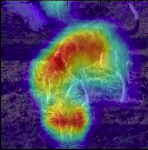}}\\

      {\includegraphics[width=2.4cm]{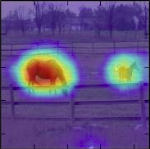} }\\

  {\includegraphics[width=2.4cm]{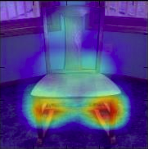} }\\

  {\includegraphics[width=2.4cm]{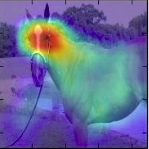} }\\

  {\includegraphics[width=2.4cm]{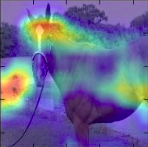} }\\
  
    {\includegraphics[width=2.4cm]{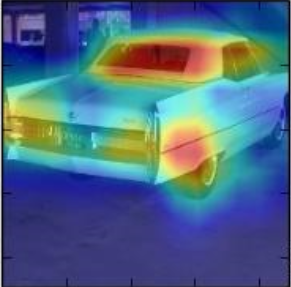} }\\

    {\includegraphics[width=2.4cm]{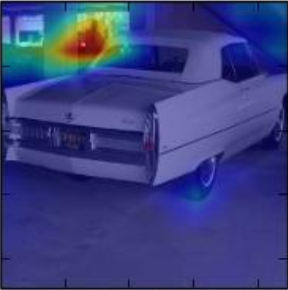} }\\

    \centerline{\textbf{Expansive}}
    \centerline{\textbf{attention map}}

\end{minipage}
%cls map**************************************************
\begin{minipage}[t]{2.6cm}
  \centering

  {\includegraphics[width=2.4cm]{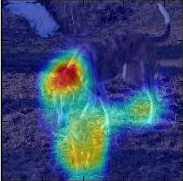}}\\

      {\includegraphics[width=2.4cm]{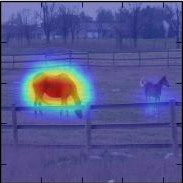} }\\

  {\includegraphics[width=2.4cm]{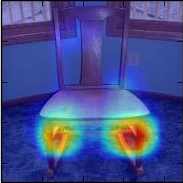} }\\

  {\includegraphics[width=2.4cm]{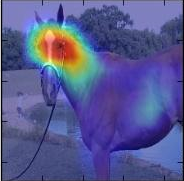} }\\

  {\includegraphics[width=2.4cm]{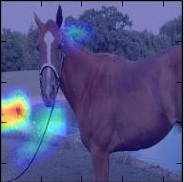} }\\
  
      {\includegraphics[width=2.4cm]{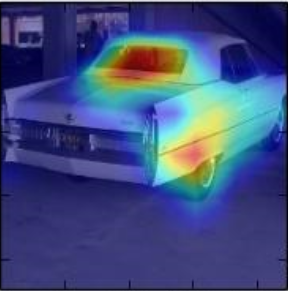} }\\

    {\includegraphics[width=2.4cm]{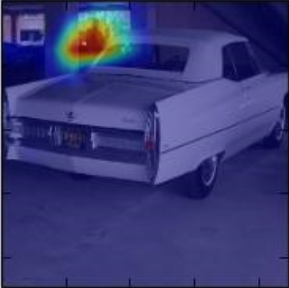} }\\

    \centerline{\textbf{Discriminative}}
    \centerline{\textbf{attention map}}
\end{minipage}
%merge map*********************************************
\begin{minipage}[t]{2.6cm}
  \centering
  {\includegraphics[width=2.4cm]{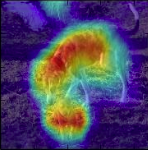}}\\

        {\includegraphics[width=2.4cm]{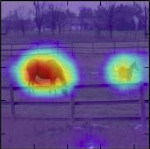} }\\

  {\includegraphics[width=2.4cm]{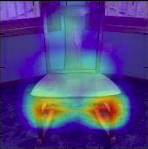} }\\

  {\includegraphics[width=2.4cm]{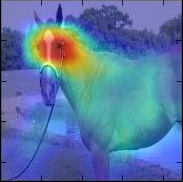} }\\

  {\includegraphics[width=2.4cm]{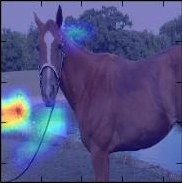} }\\
  
      {\includegraphics[width=2.4cm]{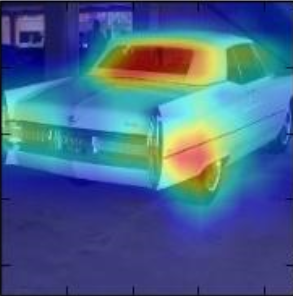} }\\

    {\includegraphics[width=2.4cm]{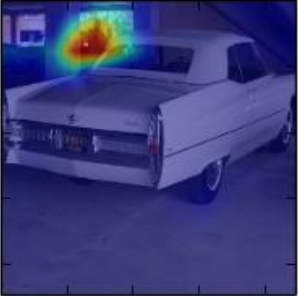} }\\

      \centerline{\textbf{merged}}
      \centerline{\textbf{attention map}}
\end{minipage}
%merge mask*************************************************
\begin{minipage}[t]{2.6cm}
  \centering
  {\includegraphics[width=2.4cm]{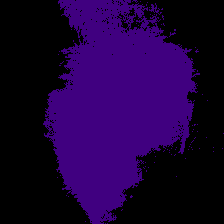}}\\

    {\includegraphics[width=2.4cm]{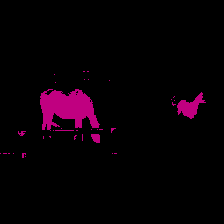} }\\

  {\includegraphics[width=2.4cm]{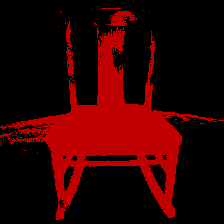} }\\

  {\includegraphics[width=2.4cm]{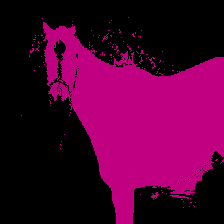} }\\

   {\includegraphics[width=2.4cm]{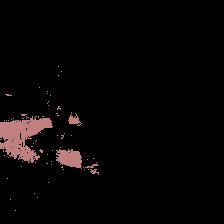} }\\

    {\includegraphics[width=2.4cm]{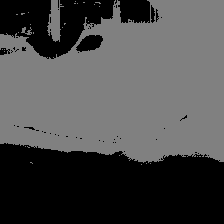} }\\

   {\includegraphics[width=2.4cm]{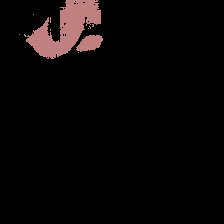} }\\

    \centerline{\textbf{pseudo-}}
    \centerline{\textbf{annotation}}
\end{minipage}
%groundtruth***********************************************
\begin{minipage}[t]{2.6cm}
  \centering
  {\includegraphics[width=2.4cm]{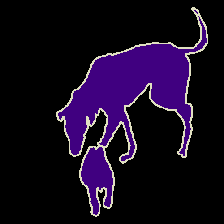}}\\

    {\includegraphics[width=2.4cm]{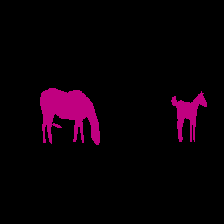} }\\

  {\includegraphics[width=2.4cm]{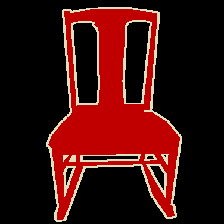} }\\

  {\includegraphics[width=2.4cm]{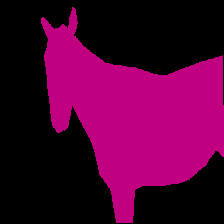} }\\
  
    {\includegraphics[width=2.4cm]{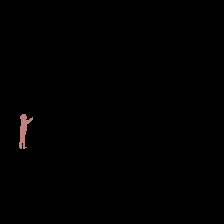} }\\

  {\includegraphics[width=2.4cm]{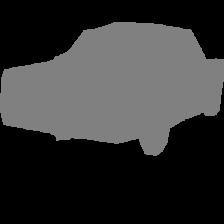} }\\
  
    {\includegraphics[width=2.4cm]{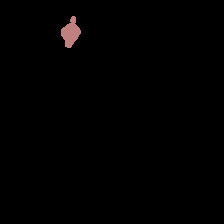} }\\

   \centerline{\textbf{ground-truth}}
   \centerline{\textbf{annotation}}
\end{minipage}

\caption{Visualization of attention maps. The first column shows the input image. The second  column shows the Expansive attention maps.  The third  column shows the Discriminative attention maps. The fourth column shows the merged attention maps. The fifth column shows the pseudo-annotations generated on the merged attention maps. The last column is the real ground-truth annotation for comparison. For the images with multiple classes we show the attention maps separately for each class. We observe that our merged attention maps combine the complimentary information of Expansive attention map and Discriminative attention maps and the generated pseudo-annotations are close to the groundtruth.}

\label{fig:Visual_all}
\end{figure*}

As described in Sec.~\ref{subsec:dual_path_atten}, our decoupled attention model outputs Expansive attention map to identify object regions and Discriminative attention map to localize the discriminative parts. We qualitatively and quantitatively compare these two attention maps generated by our decoupled attention structure and demonstrate their differences.

First we provide visual examples of the Expansive attention maps  and Discriminative attention maps of \textit{train} images in Fig.~\ref{fig:Visual_all} (column 2 and column 3). In Fig.~\ref{fig:Visual_all} we provide some examples including images with single/multiple objects and single/multiple class labels. 
We observed that for simple cases (\textit{e.g.}, large objects),  Expansive attention map is likely to cover whole region of objects, while Discriminative attention map  is likely to locate the most discriminative part. We also observed that for difficult cases (\textit{e.g.}, small objects), Expansive attention map is likely to identify a broad region, while Discriminative attention map is likely to precisely localize the object. Thus we draw the conclusion that these two attention maps have different properties and  are suitable for different situations. They are potentially complimentary to each other. Thus the combination of the two maps will result in  attention maps that are applicable to different cases which lead to object annotations of high quality.

We also quantitatively compare these two attention maps. We apply our decoupled attention model on \textit{val} images and generate annotation masks without denseCRF refinement for Expansive attention map and Discriminative attention map. We evaluate the generated estimated masks regarding to the groundtruth of \textit{val} images. 

We propose three evaluation measures for the estimated masks to demonstrate the different properties of the two attention maps:\\
1) We use the commonly used IoU to evaluate the performance of identifying the whole objects, which are denoted as:
\begin{equation}
\label{eq:expan_criteria}
S_{IoU} = \dfrac{true\,pos.}{true\, pos. + false\, pos. + false\,neg.  }
\end{equation}
2) We  propose a criteria to evaluate the localization precision within the object region, which are denoted as:
\begin{equation}
\label{eq:loc_criteria}
S_{prec} = \dfrac{true\, pos.}{true\, pos. + false\, pos. }
\end{equation}
3) We  propose a criteria to evaluate the  localization recall over the object region, which are denoted as:
\begin{equation}
\label{eq:rec_criteria}
S_{rec} = \dfrac{true\, pos.}{true\, pos. + false\, neg. }
\end{equation}
$S_{prec}$ emphasizes localizing the concentrate and small regions within the objects which not necessarily cover the whole object regions. $S_{rec}$ emphasizes the expansion of the highlighted regions by measuring whether it includes the whole range of the objects which not necessarily localize within object regions. $S_{IoU}$ emphasizes the accuracy of identifying the whole object regions which is the final criteria to indicate the quality of the attention map for generating pseudo-annotations. We use the average score over all classes, which are denoted as $mS_{IoU}$, $mS_{prec}$ and $mS_{rec}$ as our criteria.

The evaluation results are shown in the first two columns of Table~\ref{table:num}. We observe that Expansive attention map gets higher  $mS_{rec}$ and $mS_{IoU}$ score, while Discriminative attention map gets higher $mS_{prec}$. 
This verifies different properties of attention maps: Discriminative attention map is likely to localize partial interior object regions while Expansive attention map highlights the regions of larger expansion and  is likely to identify the whole object region.

As described in Sec.~\ref{subsec:dual_path_atten}, we can merge these two attention maps to improve the quality of generated pseudo-annotations. We merge the two attention maps as Eq.~\eqref{eq:merge} and follow the $mS_{IoU}$ criteria. The results are shown in the last column of Table~\ref{table:num}. We observe that $mS_{IoU}$ of merged attention map is higher than only using single type of attention map, which demonstrates that Expansive attention map and Discriminative attention map are complimentary to each other in localizing the whole object regions. The pseudo-annotations generated from the merged attention map are relatively close to the ground-truth, as is shown in Fig.~\ref{fig:Visual_all}.

\begin{table}[!htb]
\begin{center}
\caption{Evaluation of different drop-out rates (DR). With the increase of drop-out rates, $mS_{rec}$ constantly increases while  $mS_{prec}$ constantly decreases. It shows that large drop-out rates enhance the expansion effect of attention maps to some extent.}
\label{table:Dropout}
\resizebox{0.9\linewidth}{!}{
\begin{tabular}{|l|c|c|c|c|c|c|}
\hline
 DR & 0 & 0.3 & 0.4 & 0.5 & 0.6 & 0.7\\
\hline
 $mS_{prec}$ & \textbf{57.3} & 57.1 & 56.7 & 56.2 & 55.1 & 51.8\\
 
 $mS_{rec}$ & 64.5 & 65.7 & 67.6 & 69.6 & 71.6 & \textbf{74.1}\\

 $mS_{IoU}$  & 42.6 & 43.1 & 43.7 & 44.0 & \textbf{44.2} & 42.8\\
\hline
\end{tabular}
}
\end{center}
\end{table}

\subsubsection{Evaluation of dropout layers}

\begin{figure*}[htb]

\begin{minipage}[b]{0.10\linewidth}
 \centering
  \centerline{\includegraphics[width=2.1cm]{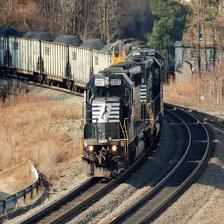}}
 \centerline{\includegraphics[width=2.1cm]{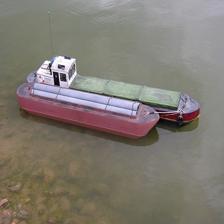}}
\centerline{input image}
\end{minipage}
 \hfill
\begin{minipage}[b]{0.10\linewidth}
 \centering
 \centerline{\includegraphics[width=2.1cm]{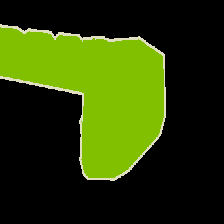}}
 \centerline{\includegraphics[width=2.1cm]{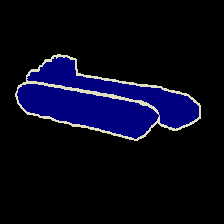}}
\centerline{ground-truth}
\end{minipage} 
\hfill
\begin{minipage}[b]{0.10\linewidth}
 \centering
 \centerline{\includegraphics[width=2.1cm]{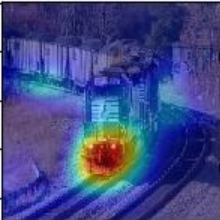}}
 \centerline{\includegraphics[width=2.1cm]{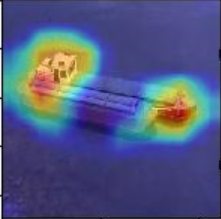}}
\centerline{DR=0.3 }

\end{minipage}
 \hfill
\begin{minipage}[b]{0.10\linewidth}
 \centering
  \centerline{\includegraphics[width=2.1cm]{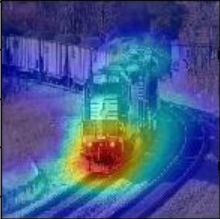}}
 \centerline{\includegraphics[width=2.1cm]{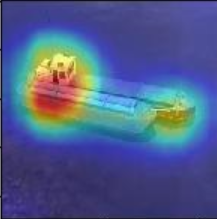}}
\centerline{DR=0.7 }
\end{minipage} 
\hfill\vline\hfill
\begin{minipage}[b]{0.10\linewidth}
 \centering
   \centerline{\includegraphics[width=2.1cm]{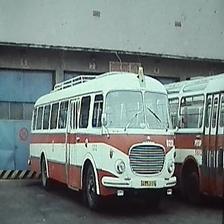}}
 \centerline{\includegraphics[width=2.1cm]{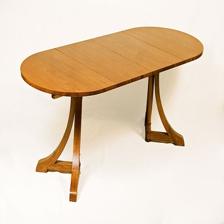}}
\centerline{input image}
\end{minipage}
 \hfill
\begin{minipage}[b]{0.10\linewidth}
 \centering
  \centerline{\includegraphics[width=2.1cm]{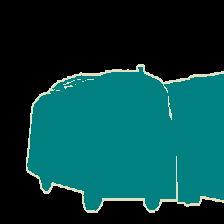}}
 \centerline{\includegraphics[width=2.1cm]{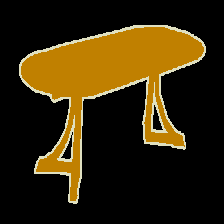}}
\centerline{ground-truth}
\end{minipage} 
\hfill
\begin{minipage}[b]{0.10\linewidth}
 \centering
  \centerline{\includegraphics[width=2.1cm]{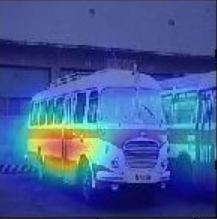}}
 \centerline{\includegraphics[width=2.1cm]{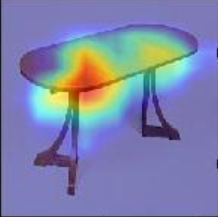}}
\centerline{DR=0.3 }

\end{minipage}
 \hfill
\begin{minipage}[b]{0.10\linewidth}
 \centering
   \centerline{\includegraphics[width=2.1cm]{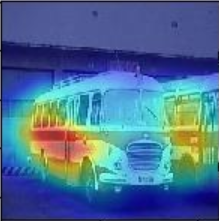}}
 \centerline{\includegraphics[width=2.1cm]{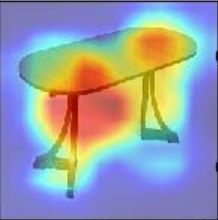}}
\centerline{DR=0.7 }
\end{minipage}

\caption{The visual examples of different drop-out rates (DR). It shows that using larger drop-out rates (DR = 0.7) leads to expansion effect over the small drop-out rates (DR = 0.3).} 
\label{fig:dropout-example}
\end{figure*}

As described in Sec.~\ref{subsec:dual_path_atten}, we add dropout layers in the Expansive attention detector modular to have expansion effect. In this section we evaluate and discuss the effect of drop-out layers.

In our experiments we use the default drop-out rate of 0.5. We also evaluate other drop-out rates and follow the same criteria as Table~\ref{table:num} on the merged attention maps. The results are reported in Table~\ref{table:Dropout}. It shows that with the increase of drop-out rates, $mS_{rec}$ constantly increases while  $mS_{prec}$ constantly decreases. This demonstrates that increasing the drop-out rates helps expand the identified region from the interior parts to the entire object scales. We also show the visualization examples of the attention maps generated by different drop-out rates in Fig.~\ref{fig:dropout-example}, which demonstrate the expansion effect results from larger drop-out rates.

The effects of drop-out layers could be explained as follows: Drop-out layers randomly set the feature grids to zeros by a rate which consequently suppress the feature space by noise. Such zero-out process is operated directly on the CNN feature  space at random  spatial locations and feature dimensions. If the  discriminative features are suppressed by drop-out process, the training process will mine other discriminative features for classification purpose. Thus the attention model will adapt to highlight more relevant features instead of the most discriminative ones.

\subsubsection{Evaluation of decoupled structure}

In this section, we aim to show that our decoupled attention structure is effective.
We implement the case of removing the Expansive attention detector modular and only train the remaining Discriminative attention detector modular. We add one drop-out layer before and after the convolutional layer of Discriminative
attention modular respectively. 
Then we evaluate the Discriminative attention map follow the same criteria as Table~\ref{table:num}. The result  of $mS_{prec}$, $mS_{rec}$ and $mS_{IoU}$ are  48.3$\%$ , 49.7$\%$ and 31.5$\%$  respectively. These results are far lower than our decoupled attention in Table~\ref{table:num}. 
It verifies that our decoupled attention structure is more effective than the single-stream attention.

\begin{table}[t]
\begin{center}
\caption{Comparisons with  Region-Mining Based Methods.  We list the middle-step segmentation results based on the pseudo-annotations generated by different region-mining based methods.  We outperform
SEC and AE w/o PSL with various settings.}  
\label{table:SEG1}
\resizebox{0.8\linewidth}{!}{

\begin{tabular}{|l|l|}
\hline
Methods & mIoU \\
\hline
\textbf{SEC} \cite{kolesnikov2016seed} & \\
$L_{seed}$ & 45.4\\
$L_{seed}+L_{constrain}$  & 50.4\\
$L_{seed}+L_{constrain}+L_{expand}$  & 50.7\\
\hline
\textbf{AE w/o PSL} \cite{wei2017object}  & \\
AE-step1  & 44.9\\
AE-step2 & 49.5\\
AE-step3 & 50.9\\
AE-step4 & 48.8\\

\hline
\textbf{Ours} & \\

Expansive attention map & 54.1\\
Discriminatve attention map & 52.2\\
merged attention map & 55.4\\

\hline
\end{tabular}
}
\end{center}
\end{table}

\subsection{Comparisons with  Region-Mining Based Methods}

Our method is related to region-mining based weakly supervised semantic segmentation approaches. In this section, we compare with two recently proposed region-mining based approaches, \textit{i.e.}, SEC \cite{kolesnikov2016seed} and Adversarial-Erasing (AE) \cite{wei2017object}. Both SEC and AE use region-mining methods CAM \cite{zhou2016learning} to generate pseudo-annotations. CAM \cite{zhou2016learning} is a method to mine the discriminate object regions by image-level labels, which is related to our attention based methods.

SEC \cite{kolesnikov2016seed} contains three losses to optimize: 
 $L_{seed}$ is the segmentation loss based on pseudo-annotations.  
 $L_{expand}$ is  for expanding the localization seed to object scale the  and $L_{constrain}$ is to make the segmentation results agree with the denseCRF \cite{hariharan2011semantic} result. 
 AE \cite{wei2017object} iteratively mine the object regions through multiple iteration steps by erasing the most discriminative regions from the original images and re-train the CAM localization network to mine the next most discriminative regions. The mined regions of all the iterations are combined as the final pseudo-annotations. We list segmentation results of the middle steps of SEC and AE in Table~\ref{table:SEG1}. For SEC, we show the segmentation results with different combination of losses. For AE, we list the segmentation results using different number of erasing steps without  prohibitive segmentation learning (PSL).
 
 For our methods, we list the segmentation results using the pseudo-annotations generated by Expansive attention map, Dircriminative attention map and merged attention map in Table~\ref{table:SEG1}. Since merged attention map achieves better performance over the other two attention maps, we will use the segmentation results generated by merged attention map for further experiments by default. 
 We outperform SEC and AE w/o PSL with various settings, which indicates our attention based region mining approach are significantly more effective than the CAM based mining methods. Moreover, our method employed a simple pipeline without complicated iterative training in contrast with AE.

\begin{table}[!t]
\begin{center}
\caption{Weakly supervised semantic segmentation result on \textit{val}  images. Our methods achieves the state-of-the-art results in the weakly supervised semantic segmentation supervised only with image labels without external sources  }
\label{table:SEG1_val}

\setlength{\tabcolsep}{10pt}
\resizebox{0.9\linewidth}{!}{

\begin{tabular}{|lcp{2.7cm}|}
\hline
Methods &  \textit{val} &    Comments\\
\hline
\textbf{Interactive Input} & &  Annotation\\
ScribbleSup\cite{lin2016scribblesup} & 63.1 &  scribble annotation\\
ScribbleSup(point)\cite{lin2016scribblesup} & 51.6 &  point annotation\\
What't the point \cite{bearman2016s} & 46.1 &  point annotation\\
BoxSup \cite{dai2015boxsup} & 62.0 &  Box annotation\\

\hline
\textbf{Image-level labels } & &  Additional Info\\
\textbf{with external source } &  & \\

STC \cite{wei2015stc} & 49.3 & {Web images}\\
Co-segmentation \cite{shen2017weakly} & 56.4 &  {Web images}\\
Webly-supervised \cite{jin2017webly} & 53.4 &  {Web images}\\
Crawled-Video \cite{hong2017weakly} & 58.1 & {Web videos}\\
%\hline
TransferNet \cite{hong2015learning} & 52.1 &  { MSCOCO  \cite{lin2014microsoft} pixelwise label}\\
AF-MCG \cite{qi2016augmented} & 54.3 &  {MCG proposal}\\
Joon \textit{et al.} \cite{oh2017exploiting} & 55.7 &  {Supervised Saliency}\\
DCSP-VGG16 \cite{chaudhry2017discovering} & 58.6 &  {Supervised Saliency}\\
DCSP-ResNet-101 \cite{chaudhry2017discovering} & 60.8 & {Supervised Saliency}\\
Mining-pixels \cite{hou2016mining} & 58.7 & {Supervised Saliency}\\

AE w/o PSL \cite{wei2017object} & 50.9 &  {Supervised Saliency}\\
AE-PSL \cite{wei2017object} & 55.0 &  {Supervised Saliency}\\
\hline
\hline
\textbf{Image-level label} & &  \\
\textbf{w/o external source }  & &  \\
MIL-FCN \cite{pathak2014fully} & 25.7 & \\
EM-Adapt \cite{papandreou2015weakly} & 38.2 & \\
BFBP \cite{saleh2016built} & 46.6 &  \\
DCSM \cite{shimoda2016distinct} & 44.1 &  \\
SEC \cite{kolesnikov2016seed} & 50.7 &  \\
AF-SS \cite{qi2016augmented} & 52.6 &  \\
Two-phase \cite{kim2017two} & 53.1 &  \\
Anirban  \textit{et al.} \cite{roy2017combining} & 52.8 &  \\
%\textbf{Ours } & & \\
\textbf{Ours-VGG16} & \textbf{55.4} &  \\
\textbf{Ours-ResNet-101} & \textbf{58.2} & \\
\hline
\end{tabular}
}
\end{center}
\end{table}

\begin{table}[!t]
\begin{center}
\caption{Weakly supervised semantic segmentation result on  \textit{test} images. Our methods achieves the state-of-the-art results in the weakly supervised semantic segmentation supervised only with image labels without external sources  }
\label{table:SEG1_test}

\setlength{\tabcolsep}{10pt}
\resizebox{0.9\linewidth}{!}{

\begin{tabular}{|lcp{2.7cm}|}
\hline
Methods &     \textit{test} & Comments\\
\hline
\textbf{Interactive Input} &  & Annotation\\
BoxSup \cite{dai2015boxsup} &  64.2 & Box annotation\\

\hline
\textbf{Image-level labels } &  & Additional Info\\
\textbf{with external source } &  & \\

STC \cite{wei2015stc} &  51.2 &{Web images}\\
Co-segmentation \cite{shen2017weakly} &  56.9 & {Web images}\\
Webly-supervised \cite{jin2017webly} &  55.3 & {Web images}\\
Crawled-Video \cite{hong2017weakly} &  58.7 & {Web videos}\\
TransferNet \cite{hong2015learning} &  51.2 & { MSCOCO  \cite{lin2014microsoft} pixelwise label}\\
AF-MCG \cite{qi2016augmented} &  55.5 & {MCG proposal}\\
Joon \textit{et al.} \cite{oh2017exploiting} &  56.7 & {Supervised Saliency}\\
DCSP-VGG16 \cite{chaudhry2017discovering} &  59.2 & {Supervised Saliency}\\
DCSP-ResNet-101 \cite{chaudhry2017discovering} &  61.9 & {Supervised Saliency}\\
Mining-pixels \cite{hou2016mining} & 59.6 & {Supervised Saliency}\\
AE w/o PSL \cite{wei2017object} &  & {Supervised Saliency}\\
AE-PSL \cite{wei2017object} &  55.7 & {Supervised Saliency}\\
\hline
\hline
\textbf{Image-level label} &  & \\
\textbf{w/o external source }   & & \\
MIL-FCN \cite{pathak2014fully}  & 24.9 &\\
EM-Adapt \cite{papandreou2015weakly}  & 39.6 &\\
BFBP \cite{saleh2016built} &  48.0 & \\
DCSM \cite{shimoda2016distinct} &  45.1 & \\
SEC \cite{kolesnikov2016seed} &  51.7 & \\
AF-SS \cite{qi2016augmented} &  52.7 & \\
Two-phase \cite{kim2017two} &  53.8 & \\
Anirban  \textit{et al.} \cite{roy2017combining} &  53.7 & \\
\textbf{Ours-VGG16} &  \textbf{56.4} & \\
\textbf{Ours-ResNet-101} &  \textbf{60.1} &\\
\hline
\end{tabular}
}
\end{center}
\end{table}

\subsection{Comparisons with State-of-the-arts}

\setlength{\tabcolsep}{3pt}
\begin{table*}[!tp]
\begin{center}
\caption{Our Segmentation results for each class on \textit{val} and \textit{test} images. We list the results of using Deeplab-largeFOV (vgg16) \cite{chen2014semantic} and ResNet-101 based Deeplab-like model (resnet) \cite{shen2017learning} as segmentation DCNN model. }
\label{table:Full_result}
%\normalsize
\resizebox{0.95\linewidth}{!}{
\begin{tabular}{|l|ccccccccccccccccccccc|l|}
\hline
\noalign{\smallskip}
& \rotatebox{90}{background}& \rotatebox{90}{aeroplane}&\rotatebox{90}{bicycle}&\rotatebox{90}{bird}&\rotatebox{90}{boat}&\rotatebox{90}{bottle}&\rotatebox{90}{bus}&\rotatebox{90}{car}&\rotatebox{90}{cat}&\rotatebox{90}{chair}&\rotatebox{90}{cow} &\rotatebox{90}{dining table}&\rotatebox{90}{dog}&\rotatebox{90}{horse}&\rotatebox{90}{motorbike}&\rotatebox{90}{person}&\rotatebox{90}{potted plant}&\rotatebox{90}{sheep}&\rotatebox{90}{sofa}&\rotatebox{90}{train}&\rotatebox{90}{tv/monitor}&\rotatebox{90}{mIoU}\\

\hline

vgg16-\textit{val}& 84.5 & 73.3 & 23.4 & 
66.4 & 42.7&  48.7 & 76.9 & 61.0 & 66.0 & 21.8 & 65.2 & 37.1 &70.0 &  62.5 & 62.0 & 58.9 & 41.1 & 71.2 & 32.4 & 58.7 & 39.0 & 55.4 \\

\hline

vgg16-\textit{test}& 85.6 & 71.2 & 27.1 & 
74.2 & 35.8& 51.0& 77.7 & 63.4 & 65.0 & 21.4 & 60.0 & 47.7 & 73.9 & 60.2& 71.0 & 61.5 & 39.2 & 64.0 & 39.5 & 54.2 & 40.0 & 56.4\\

\hline

resnet-\textit{val} & 84.7 & 73.5 & 27.1 & 
68.3 &  43.8   & 57.2 & 80.9 & 64.7 & 65.4 & 22.3 & 69.7 & 39.2 & 70.8 & 73.0 & 67.8 & 60.0 & 45.8 & 72.6 & 35.9 & 57.7 & 42.2 & 58.2\\

\hline

resnet-\textit{test} & 86.3 & 70.4 & 29.7 & 
78.8 &  40.8 & 53.9  & 80.6 & 67.7 & 67.7 & 21.0 & 70.2 & 46.4 & 74.8 & 70.6 & 74.8 & 63.7 & 45.2 & 76.9 & 45.0 & 57.8 & 39.8 & 60.1 \\

\hline
\end{tabular}
}
\end{center}
\end{table*}

\begin{figure*}[!htb]

\begin{minipage}[b]{0.11\linewidth}
 \centering
  \centerline{\includegraphics[width=2.0cm]{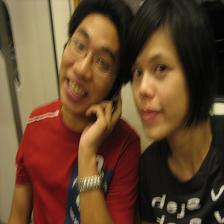}}
 \centerline{\includegraphics[width=2.0cm]{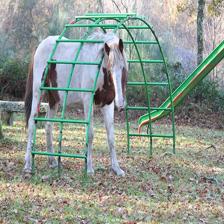}}
   \centerline{\includegraphics[width=2.0cm]{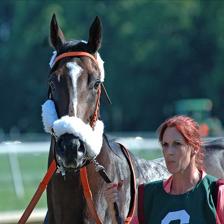}}
 \centerline{\includegraphics[width=2.0cm]{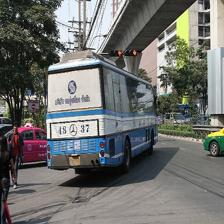}}
  \centerline{\includegraphics[width=2.0cm]{figures/JPEGImages_select/2007_006761.jpg}}
  \centerline{\includegraphics[width=2.0cm]{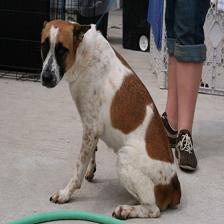}}
  \centerline{\includegraphics[width=2.0cm]{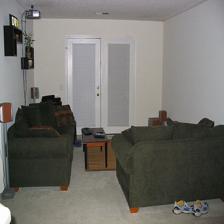}}
  \centerline{\includegraphics[width=2.0cm]{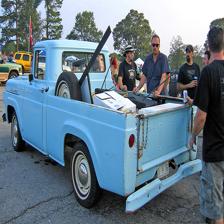}}

\centerline{input image}
\end{minipage}
 \hfill
\begin{minipage}[b]{0.11\linewidth}
 \centering
 \centerline{\includegraphics[width=2.0cm]{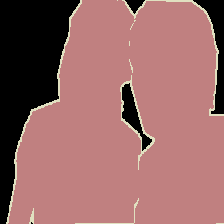}}
 \centerline{\includegraphics[width=2.0cm]{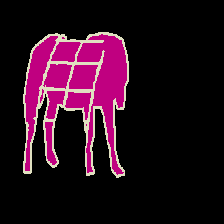}}
  \centerline{\includegraphics[width=2.0cm]{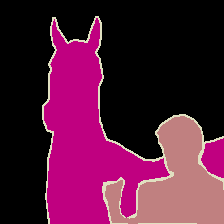}}
 \centerline{\includegraphics[width=2.0cm]{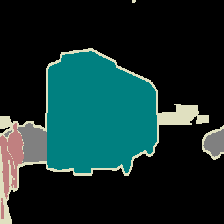}}
  \centerline{\includegraphics[width=2.0cm]{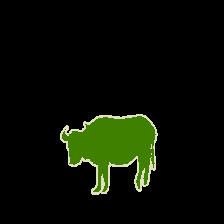}}
  \centerline{\includegraphics[width=2.0cm]{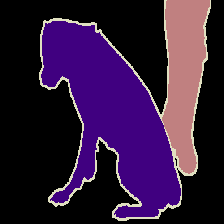}}
  \centerline{\includegraphics[width=2.0cm]{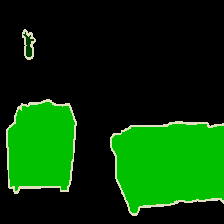}}
  \centerline{\includegraphics[width=2.0cm]{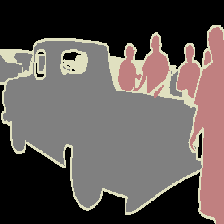}}

\centerline{ground-truth}
\end{minipage} 
\hfill
\begin{minipage}[b]{0.11\linewidth}
 \centering
 \centerline{\includegraphics[width=2.0cm]{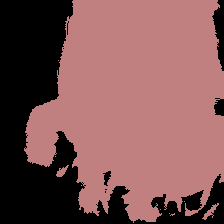}}
 \centerline{\includegraphics[width=2.0cm]{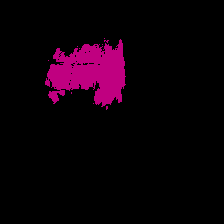}}
  \centerline{\includegraphics[width=2.0cm]{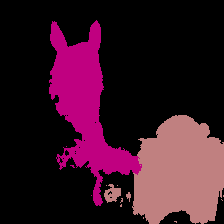}}
 \centerline{\includegraphics[width=2.0cm]{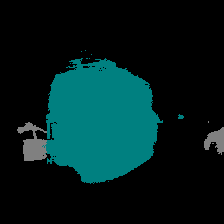}}
  \centerline{\includegraphics[width=2.0cm]{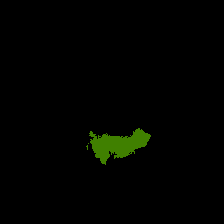}}
  \centerline{\includegraphics[width=2.0cm]{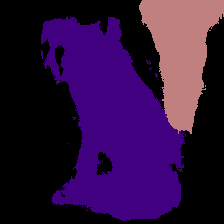}}
  \centerline{\includegraphics[width=2.0cm]{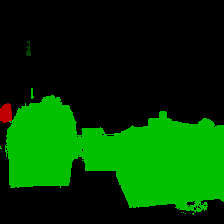}}
  \centerline{\includegraphics[width=2.0cm]{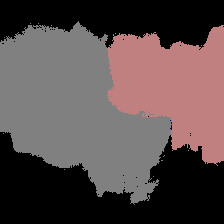}}

\centerline{Ours-VGG16}
\end{minipage}
 \hfill
\begin{minipage}[b]{0.11\linewidth}
 \centering
  \centerline{\includegraphics[width=2.0cm]{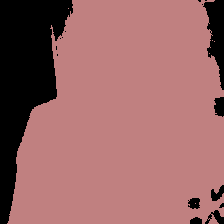}}
 \centerline{\includegraphics[width=2.0cm]{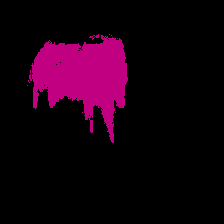}}
   \centerline{\includegraphics[width=2.0cm]{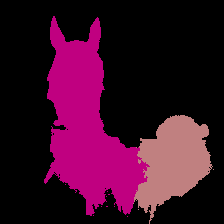}}
 \centerline{\includegraphics[width=2.0cm]{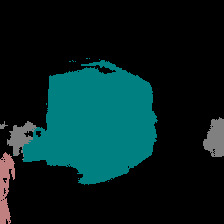}}
  \centerline{\includegraphics[width=2.0cm]{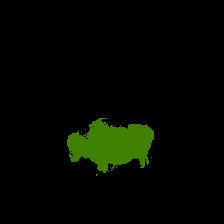}}
  \centerline{\includegraphics[width=2.0cm]{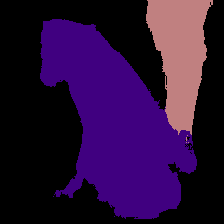}}
  \centerline{\includegraphics[width=2.0cm]{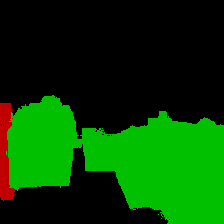}}
  \centerline{\includegraphics[width=2.0cm]{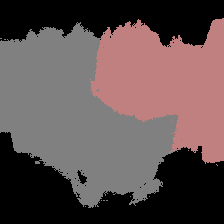}}

\centerline{Ours-ResNet-101}
\end{minipage} 
 \hfill
\begin{minipage}[b]{0.11\linewidth}
 \centering
  \centerline{\includegraphics[width=2.0cm]{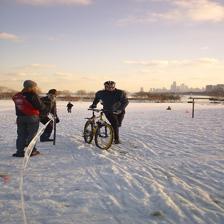}}
 \centerline{\includegraphics[width=2.0cm]{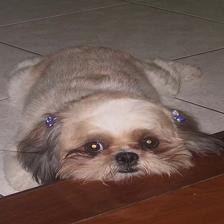}}
   \centerline{\includegraphics[width=2.0cm]{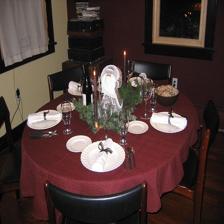}}
 \centerline{\includegraphics[width=2.0cm]{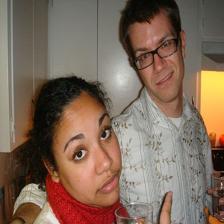}}
  \centerline{\includegraphics[width=2.0cm]{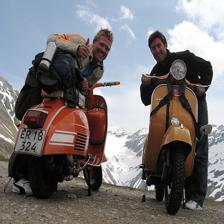}}
  \centerline{\includegraphics[width=2.0cm]{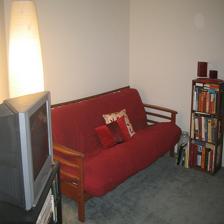}}
  \centerline{\includegraphics[width=2.0cm]{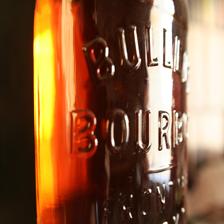}}
  \centerline{\includegraphics[width=2.0cm]{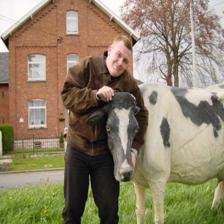}}

\centerline{input image}
\end{minipage}
 \hfill
\begin{minipage}[b]{0.11\linewidth}
 \centering
 \centerline{\includegraphics[width=2.0cm]{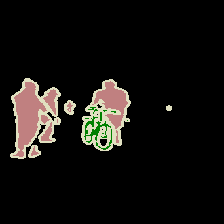}}
 \centerline{\includegraphics[width=2.0cm]{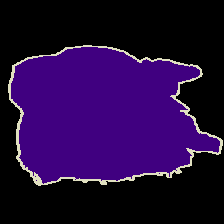}}
  \centerline{\includegraphics[width=2.0cm]{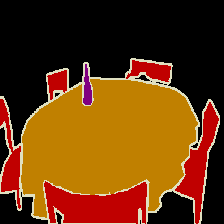}}
 \centerline{\includegraphics[width=2.0cm]{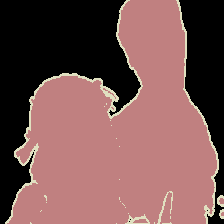}}
  \centerline{\includegraphics[width=2.0cm]{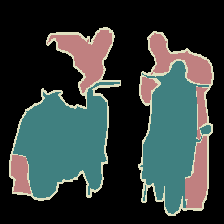}}
  \centerline{\includegraphics[width=2.0cm]{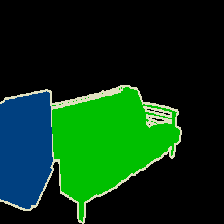}}
  \centerline{\includegraphics[width=2.0cm]{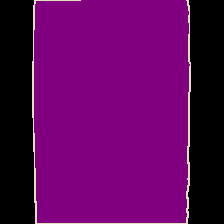}}
  \centerline{\includegraphics[width=2.0cm]{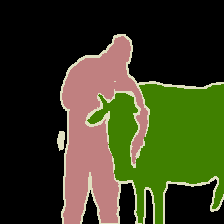}}

\centerline{ground-truth}
\end{minipage} 
\hfill
\begin{minipage}[b]{0.11\linewidth}
 \centering
 \centerline{\includegraphics[width=2.0cm]{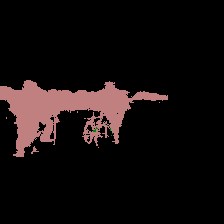}}
 \centerline{\includegraphics[width=2.0cm]{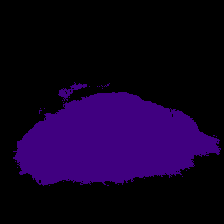}}
  \centerline{\includegraphics[width=2.0cm]{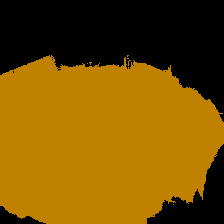}}
 \centerline{\includegraphics[width=2.0cm]{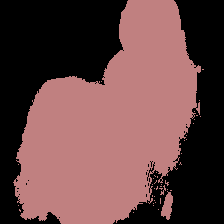}}
  \centerline{\includegraphics[width=2.0cm]{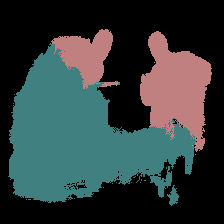}}
  \centerline{\includegraphics[width=2.0cm]{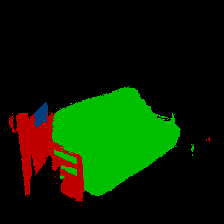}}
  \centerline{\includegraphics[width=2.0cm]{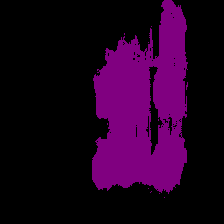}}
  \centerline{\includegraphics[width=2.0cm]{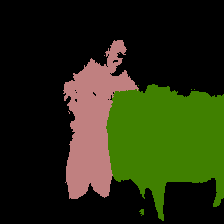}}

\centerline{Ours-VGG16}
\end{minipage}
 \hfill
\begin{minipage}[b]{0.11\linewidth}
 \centering
 \centerline{\includegraphics[width=2.0cm]{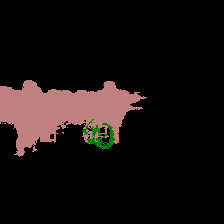}}
 \centerline{\includegraphics[width=2.0cm]{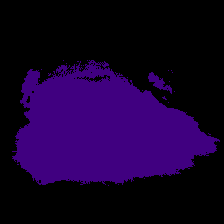}}
  \centerline{\includegraphics[width=2.0cm]{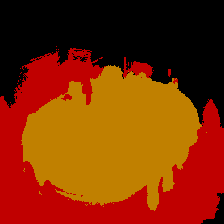}}
 \centerline{\includegraphics[width=2.0cm]{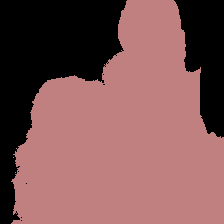}}
  \centerline{\includegraphics[width=2.0cm]{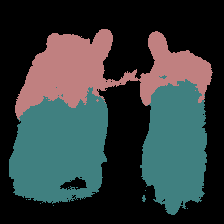}}
  \centerline{\includegraphics[width=2.0cm]{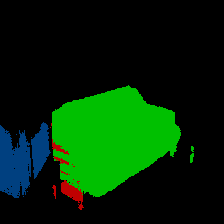}}
  \centerline{\includegraphics[width=2.0cm]{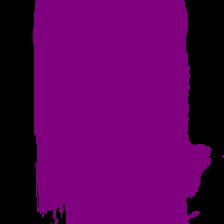}}
  \centerline{\includegraphics[width=2.0cm]{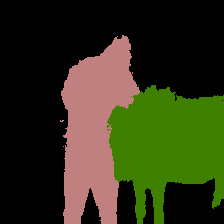}}

\centerline{Ours-ResNet-101}
\end{minipage}

\caption{ Qualitative results of PASCAL 2012 val images. We list the qualitative examples generated by Deeplab-largeFOV (Ours-VGG16) segmentation model and Resnet101 based (Ours-Resnet-101) \cite{shen2017learning} Deeplab-like segmentation model.}
\label{fig:visual_results}
\end{figure*}

In this section, we compare our segmentation results with other weakly supervised semantic segmentation methods. The comparison results are listed in Table~\ref{table:SEG1_val} and Table~\ref{table:SEG1_test} for \textit{val} and \textit{test} images respectively. We divide weakly supervised semantic segmentation into three categories based on different levels of weak supervisions and whether additional information (data/supervision) is implicitly used in their pipeline:

In the first category, the methods utilize interactive input such as scribble \cite{lin2016scribblesup} or point \cite{bearman2016s} as supervision, which is relatively more precise indicator of object location and scale. The results in this category are listed in the first block of Table~\ref{table:SEG1_val} and Table~\ref{table:SEG1_test}.

In the second category, the methods only use image-level label as supervision, but introduce information of external sources to improve segmentation results. The segmentation performance in this category are listed in the second block of Table~\ref{table:SEG1_val} and Table~\ref{table:SEG1_test}. Some work use additional data to assist training. For example, STC \cite{wei2015stc} crawled 50K web images for initial training step and Crawled-Video \cite{hong2017weakly} crawled large amount of  online videos, which significantly  increase the training data amount. Some work implicitly utilize the pixel-wise annotations in their training. For example,
 TransferNet \cite{hong2015learning} transfer the pixel-wise annotation from MSCOCO dataset \cite{lin2014microsoft} to PASCAL dataset. 
AF-MCG \cite{qi2016augmented} utilize MCG proposal method which is trained with PASCAL VOC pixel-wise ground-truth. 
Some works  utilize fully-supervised saliency detection methods in localization seed expansion \cite{oh2017exploiting} or foreground/background detection \cite{wei2015stc,wei2017object} , which implicitly utilize external saliency ground-truth data to train saliency detector.

In the third category, the methods only use image-level label as supervision without the information of external sources. Our methods belong to this category. The segmentation performance in this category are listed in the third block of Table~\ref{table:SEG1_val} and Table~\ref{table:SEG1_val}.

 We mainly focus on comparing with the methods in the third category. We achieve the mIoU score of 55.4 and 56.4 on \textit{val} and \textit{test} images respectively, which significantly outperforms other methods using the same supervision settings. 
 In order to further improve our segmentation results, we use Deeplab-like model in \cite{shen2017learning} based on Resnet-101 \cite{he2016deep} as the segmentation model. We  achieve the mIoU score of 58.2 on \textit{val} images and 60.1 on \textit{test} images, which outperforms all the existing methods in weakly supervised semantic segmentation using the same level of supervision setting up to date. We also list our segmentation results for each class in the Table~\ref{table:Full_result} for reference. We show the qualitative examples of the segmentation results in Fig.~\ref{fig:visual_results}.
 
 Moreover, we also want to emphasize that we do not iteratively train the models in multiple steps. The whole pipeline only need training decoupled attention structure and segmentation structure once on the PASCAL VOC \textit{train} data. To our knowledge our approach has the most simple pipeline for weakly-supervised semantic segmentation.

\section{Conclusion}
In this work we have presented a novel decoupled spatial neural attention work to generate pseudo-annotations for weakly-supervised semantic segmentation with image-level labels. This decoupled  attention model  simultaneously outputs two class-specific attention maps with different properties and are effective for estimating pseudo-annotations.
 We perform detailed ablation experiments to verify the effectiveness of our decoupled attention structure. Finally we achieve state-of-the-art weakly supervised semantic segmentation results  on Pascal VOC 2012 image segmentation benchmark.

\section{Acknowledgement}
\label{sec:Ack}

This research was carried out at the Rapid-Rich Object Search (ROSE) Lab at the Nanyang Technological University, Singapore. The ROSE Lab is supported by the National Research Foundation, Singapore, under its IDM Futures Funding Initiative and administered by the Interactive and Digital Media Programme. We also thank NVIDIA for their donation of GPU.

\bibliographystyle{IEEEtran}
%\bibliography{refs_seg,refs_cam,refs_atten,clas}
%\bibliography{bare_jrnl_zty}

\end{document}